\pdfoutput=1
\documentclass[11pt]{article}

\usepackage[preprint]{acl}

\usepackage{times}
\usepackage{float}
\usepackage{latexsym}
\usepackage{amsmath}
\usepackage{enumitem}
\usepackage{multirow}
\usepackage{placeins}

\usepackage[T1]{fontenc}
\usepackage[utf8]{inputenc}

\usepackage{microtype}

\usepackage{inconsolata}

\usepackage{graphicx}

\title{Robustness Evaluation of OCR-based Visual Document Understanding under Multi-Modal Adversarial Attacks
}

\author{
  Dong Nguyen Tien \\
  VinUniversity \\
  Hanoi, Vietnam \\
  \texttt{25dong.nt@vinuni.edu.vn} \\
  \texttt{ntdong@cmc.com.vn} \\\And
  Dung D. Le \\
  VinUniversity \\
  Hanoi, Vietnam \\
  \texttt{dung.ld@vinuni.edu.vn} \\
}
\begin{document}
\maketitle
\begin{abstract}
Visual Document Understanding (VDU) systems have achieved strong performance in information extraction by integrating textual, layout, and visual signals. However, their robustness under realistic adversarial perturbations remains insufficiently explored. We introduce the first unified framework for generating and evaluating multi-modal adversarial attacks on OCR-based VDU models. Our method covers six gradient-based layout attack scenarios, incorporating manipulations of OCR bounding boxes, pixels, and texts across both word and line granularities, with constraints on layout perturbation budget (e.g., IoU $\geq$ 0.6) to preserve plausibility.

Experimental results across four datasets (FUNSD, CORD, SROIE, DocVQA) and six model families demonstrate that line-level attacks and compound perturbations (BBox + Pixel + Text) yield the most severe performance degradation. Projected Gradient Descent (PGD)-based BBox perturbations outperform random-shift baselines in all investigated models. Ablation studies further validate the impact of layout budget, text modification, and adversarial transferability.
\end{abstract}

%=====================================================================
\section{Introduction}
\label{sec:intro}
Recent advances in \emph{Visual Document Understanding} (VDU) have enabled automated information-extraction and question-answering pipelines for banking, taxation, legal compliance, and e-government services. Multimodal Transformers that jointly encode text, layout, and visual cues-such as LayoutLMv2 \citep{layoutlmv2}, LayoutLMv3 \citep{layoutlmv3}, DocFormer \citep{docformer}, ERNIE \citep{ernie}, and GeoLayoutLM \citep{geolayoutlm}-achieve high positions in the leaderboards on datasets like FUNSD \citep{funsd}, CORD \citep{cord}, SROIE \citep{sroie}, and DocVQA \citep{docvqa}. 

\noindent\textbf{OCR-based VDU families.}  
One can categorize the existing OCR-based VDU models into one of two following branches, based on the required input modalities:  
\begin{enumerate}[label=\arabic*)]
  \item \textbf{Text + Layout + Image:} LayoutLMv2 \citep{layoutlmv2}, LayoutLMv3 \citep{laytextllm}, ERNIE\citep{ernie} and GeoLayoutLM \citep{geolayoutlm}.
  \item \textbf{Text + Layout:} LayTextLLM \citep{laytextllm} and a prompt-engineered Llama\citep{llama3} baseline that serialises every bbox as a single token.
\end{enumerate}

\paragraph{Research Gap and Motivation.} Prior work on VDU robustness has tackled distribution shift~\citep{dogood}, image corruption~\citep{rodla}, or unicode text attacks~\citep{whenvisionfails}. However, none of the mentioned work has investigated a unified attacking framework of BBox, text, and image constraint-based perturbations. To this end, we introduce a multi-modal adversarial framework targeting OCR-based VDU models, with considerations of perturbation budgets of related modalities (BBox, text, and image).

OCR-based pipelines remain dominant in practice due to their traceability. APIs like APIs-Amazon Textract \citep{aws_bbox_doc}, Azure Document Intelligence \citep{azure_read_api}, and Google Document AI \citep{google_enterprise_docai} return bounding boxes crucial for auditability in domains like finance. OCR-free models~\citep{internvl,llavanext,qwen25vltechnicalreport} are emerging but still lag in fine-grained extraction and spatial grounding.

Our unified attacks expose severe vulnerabilities in OCR-based models (up to 29.18\% F$_1$ drop), highlighting the urgent need for robustness tools in layout-aware VDU systems.

%=====================================================================

\noindent \textbf{} We summarize our contributions as follows:

\begin{itemize}
  \item \textbf{Unified Multi-Modal Attack Framework.} We propose the first framework for adversarial attacks on VDU models across \textit{layout}, \textit{text}, and \textit{image} modalities under a shared budget. PGD-based layout attacks leverage a differentiable mIoU loss with IoU $\in \{0.6, 0.75, 0.9\}$.
  
  \item \textbf{Scenario-Based Benchmarking.} We define six attack scenarios across word- and line-levels, evaluated on four standard datasets: \textsc{FUNSD} ~\citep{funsd}, \textsc{CORD} ~\citep{cord}, \textsc{SROIE} ~\citep{sroie}, \textsc{DocVQA} ~\citep{docvqa}.
  \item \textbf{Robustness Analysis and Insights.} Our findings:
  \begin{itemize}
    \item Line-level attacks consistently outperform word-level.
    \item PGD is more effective than random shift, even under tight layout budgets.
    \item Unicode diacritic attacks cause larger degradation than random text edits.
    \item PGD attacks transfer well to models without visual input (e.g., LayTextLLM ~\citep{laytextllm}, Llama ~\citep{laytextllm}).
  \end{itemize}
\end{itemize}

%=====================================================================
\section{Related Work}
\label{sec:related_work}

\subsection{OCR-based VDU Models}
OCR‐based VDU models can be grouped by input modality as following:

\noindent\textbf{Text + Layout + Image} models including LayoutLMv2~\citep{layoutlmv2}, LayoutLMv3~\citep{layoutlmv3}, ERNIE-Layout~\citep{ernie} and GeoLayoutLM~\citep{geolayoutlm}-concatenate token embeddings with 2-D positional encodings and CNN/ViT image features.

\noindent\textbf{Text + Layout} models such as LayTextLLM~\citep{laytextllm}
and a prompt-engineered LLaMA‐3~\citep{llama3} baseline serialise each bounding box as an additional token, removing the image branch while retaining spatial structure.Although these architectures achieve state-of-the-art accuracy on FUNSD~\citep{funsd}, CORD~\citep{cord}, SROIE~\citep{sroie} and DocVQA~\citep{docvqa}, they still rely on \emph{discretised bounding-box embeddings} that can be shifted at inference time\,-a vulnerability we exploit in this work.

\subsection{Robustness and Adversarial Attacks}

\noindent\textbf{Document-specific robustness.}  
Do-GOOD~\citep{dogood} probes distribution shift \emph{but keeps the original bounding boxes intact}, leaving layout robustness unexplored. RoDLA~\citep{rodla} introduces a robustness suite consisting of 12 image perturbations applied to documents. The work of~\citep{doctamper} focuses on tampering detection and localization but does not generate adversarial examples.

\noindent\textbf{Multimodal LLM Safety.}  
Recent safety evaluations for multimodal large language models (MLLMs) expose vulnerabilities to image or prompt-based adversarial triggers. Key works include the safety benchmark for MLLMs~\citep{safety_mllm}, ImgTrojan~\citep{imgtrojan}, and adversarial jailbreaks through visual examples~\citep{visualadversarialexamplesjailbreak}. We also note ongoing work in 2025 exploring new variants of ImgTrojan and visual AE-based attacks, highlighting the lack of grounded layout-aware benchmarks.

\noindent\textbf{Text-in-image attacks.}  
\citep{whenvisionfails} present a genetic algorithm to perturb words inside images that fool OCR systems and ViT backbones, highlighting the brittleness of current OCR-based perception. These efforts underscore the need for a layout-aware, budget-controlled benchmark, which we introduce in this work.

\noindent\textbf{OCR-free vision–language models.}
InternVL~\citep{internvl}, LLaVA-1.5~\citep{llavanext}, Qwen-VL~\citep{qwen25vltechnicalreport} and DeepSeek-VL~\citep{deepseekvl} bypass explicit OCR by decoding text directly from pixels.  Because they do not expose token-level bounding boxes, their attack surface is fundamentally different from that of OCR-based models.  We therefore focus our robustness study on the still-dominant OCR-based family and leave the adaptation of our layout-budget concept to patch-level perturbations for OCR-free systems to future work (see Section~\ref{sec:intro} for a detailed motivation).

\subsection{Bounding-Box Perturbations in Vision and Document Layout}
\textbf{Generic object detection.}
Distortion-Aware BBox Attack~\citep{distortionawareadversarialattacksbounding} and
ABBG~\citep{adversarialbboxgeneration} perturb bounding boxes to mislead detectors and trackers.  
However, these methods are designed for natural images and do not
consider textual semantics or document layout constraints.  
To the best of our knowledge, \emph{no prior work systematically
evaluates adversarial robustness of OCR-based VDU models to bounding-box layout shifts}-let alone in combination with pixel and text
perturbations.  
We close this gap by proposing a budget-controlled, multi-modal attack
suite and a differentiable PGD strategy tailored to discrete layout
embeddings.

%=====================================================================

\section{Method}
\label{sec:method}

Our goal is to devise a unified \textbf{ budgeted multimodal attack framework} that enables simultaneous perturbations of \emph{layout}, \emph{text} and \emph{pixel} channels of OCR-based VDU models while keeping each perturbation within an interpretable budget.
Figure~\ref{fig:pipeline_overview} gives an overview.

%--------------------------------------------------------------------
\subsection{Threat Model and Budget Constraints}
\label{ssec:budget}

We define a unified threat model
$\mathcal{T} = (\mathcal{B}_{\text{layout}},\mathcal{B}_{\text{text}},\mathcal{B}_{\text{pixel}})$:

\begin{itemize}[leftmargin=*]
    \item \textbf{Layout budget:} $$\mathcal{B}_{\text{layout}}\!: \operatorname{IoU}(B,\tilde B)\!\ge\!\tau,\;\tau\!\in\!\{0.9,0.75,0.6\}.$$ IoU measures the overlap between the original bounding box $B$ and perturbed box $\tilde B$, ensuring layout perturbations remain spatially consistent. Adversarial bounding boxes $\tilde B$ are generated by either (i) randomly shifting the original box $B$ in one of four directions or scaling it slightly, or (ii) optimizing $\tilde B$ via PGD to minimize task loss, under the constraint of a minimum IoU.

    \item \textbf{Text budget:} $$\mathcal{B}_{\text{text}}\!: \mathrm{edit\_rate}(x,\tilde x)\!\in\!\{0,0.1\}.$$ The edit\_rate quantifies character-level replacements between clean text $x$ and adversarial text $\tilde x$, constrained to visually plausible noise. We randomly replace characters in the original text $x$ with other characters at a fixed rate (0.1 or change Unicode). We do not allow insertions or deletions, ensuring that token positions remain aligned.

    \item \textbf{Pixel budget:} $$\mathcal{B}_{\text{pixel}}\!: T\!\in\!\mathcal{T}_{\text{RoDLA}}.$$ $\mathcal{T}_{\text{RoDLA}}$ denotes a transformation methods from \texttt{RoDLA}~\citep{rodla}, a set of 12 document-specific augmentations (e.g., blur, noise, occlusion). The image region inside a bounding box is first shifted according to the adversarial box $\tilde B$, then optionally augmented using one randomly sampled transformation from the 12 document-centric visual effects defined in \citet{rodla} (e.g., blur, contrast, noise, shadow).
\end{itemize}

%--------------------------------------------------------------------
\subsection{Learning-Based Bounding-Box Reparameterisation}
\label{ssec:bbx_pred}

\paragraph{BBox Predictor.} To enable gradient-based layout attacks, we train a compact BBox predictor $g_\theta$ that maps each token embedding $\mathbf{e}_i$ to a tuple $\langle c_x, c_y, \log w, \log h \rangle$. The architecture consists of a 2-layer MLP for input projection, a 4-layer Transformer encoder, and a 2-layer output MLP. We optimize using a combined SmoothL1 and GIoU loss:
\begin{equation}
  \mathcal{L}_{\text{box}} = \mathcal{L}_{\text{SmoothL1}} + \lambda_{\text{GIoU}}\cdot \mathcal{L}_{\text{GIoU}}(\hat b, b)
  \label{eq:boxloss}
\end{equation}
\noindent where \(\lambda_{\text{GIoU}} = 2.0\).
%--------------------------------------------------------------------
\begin{figure*}[t]
  \centering
  \includegraphics[width=0.8\textwidth]{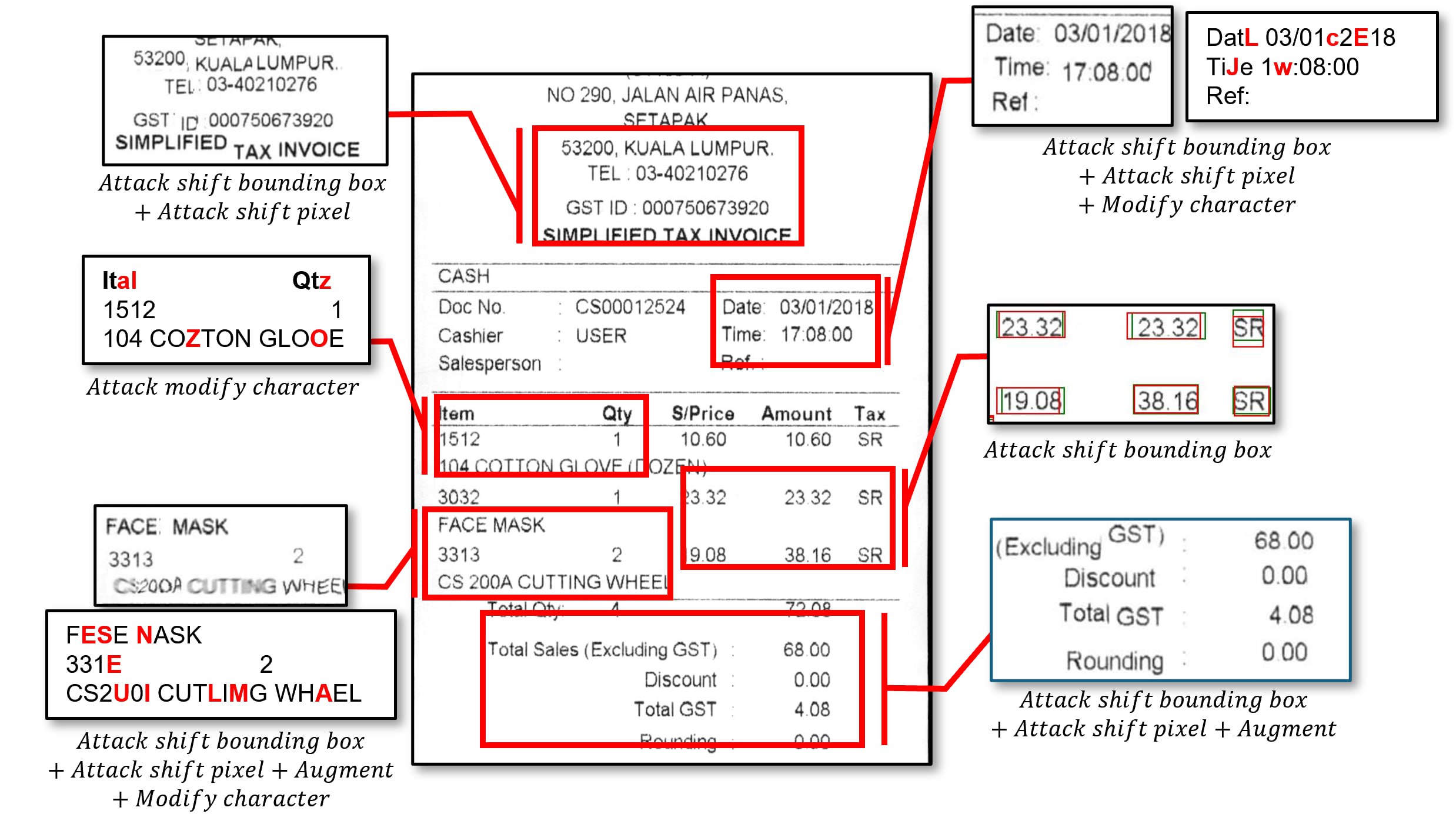}
  \caption{Overview of the six proposed adversarial attack scenarios apply in each sample of the datasets}
  \label{fig:pipeline_overview}
\end{figure*}
%--------------------------------------------------------------------

\subsection{PGD with an mIoU-Budget Loss}
\label{ssec:pgd}

Let \(\hat{\mathbf B}=g_\theta(\mathbf e)\) be the boxes predicted from the
clean token embeddings \(\mathbf e\).
To craft an adversarial embedding \(\tilde{\mathbf e}\) we
\emph{maximise}
\[
  \mathcal L_{\text{adv}}
  = \mathcal L_{\text{task}}
  - \lambda_{\text{box}}\bigl[1-\operatorname{IoU}(\hat{\mathbf B},
                                                  \tilde{\mathbf B})\bigr],
\]
where \(\tilde{\mathbf B}=g_\theta(\tilde{\mathbf e})\) and
\(\lambda_{\text{box}}\!>\!0\).
At each PGD step (\(T{=}10\), \(\alpha{=}0.05\)) we update
\(\mathbf e\) along \(\nabla_{\mathbf e}\mathcal L_{\text{adv}}\) and
\textbf{project} the resulting boxes back into the feasible set
\(\{\operatorname{IoU}\ge\tau\}\), with
\(\tau\!\in\!\{0.9,0.75,0.6\}\).
Among the ten candidates we keep the one that
maximises \(\mathcal L_{\text{task}}\) while still meeting the IoU
budget.

%--------------------------------------------------------------------
\subsection{Text and Pixel Modules}
\label{ssec:text_pixel}

\begin{itemize}
  \item \textbf{Text.} Two strategies:  
  (i) random character replacement ($\rho\!=\!0.1$), and  
  (ii) Unicode-combining genetic optimisation \citep{whenvisionfails}.

  \item \textbf{Pixel.} For any shifted box we  
  (a) translate the enclosed pixels to match the new box and optionally  
  (b) apply one RoDLA transform~\citep{rodla}, yielding visually coherent perturbations.
\end{itemize}

%--------------------------------------------------------------------
\subsection{Attack Scenarios and Granularities}
\label{ssec:scenarios}

We evaluate six scenarios using both \emph{word-level} and \emph{line-level} granularity, where word-level refers to assessing the accuracy of individual words recognized by the OCR model, while line-level considers the correctness of entire lines of text, including word order and spacing.

\begin{enumerate}[label=\textbf{S\arabic*},leftmargin=*]
  \item \textbf{BBox only}:
        shift bounding boxes; pixels and text frozen.
  \item \textbf{BBox + Pixel}:
        shift boxes and translate the same pixel region.
  \item \textbf{BBox + Pixel + Augment}:
        S2 plus one RoDLA transform ~\citep{rodla}.
  \item \textbf{Text only}:
        mutate text under $\rho$; layout and image untouched.
  \item \textbf{BBox + Text}:
        change layout and text jointly.
  \item \textbf{BBox + Pixel + Text}:
        full multi-modal attack across all three channels.
\end{enumerate}

\subsection{Pipeline Overview}
\label{ssec:pipeline}

Figure~\ref{fig:pipeline_overview} illustrates our modular attack pipeline. 
The attacker operates on OCR-based documents by modifying layout (via bounding box embeddings), text (via character-level or Unicode perturbations), and image content (via RoDLA-based pixel transformations). 
Perturbations are bounded by a unified budget, and gradient flow is enabled through a differentiable BBox-Predictor.

Given a document image and OCR output, the pipeline produces perturbed inputs that are fed into a frozen VDU model.
The resulting adversarial document is then evaluated on key information extraction (KIE) or document visual QA metrics.

%=====================================================================
\section{Experimental Results}

\subsection{Experimental Setup}
\label{ssec:exp_setup}

We evaluate our adversarial framework on four widely used benchmarks:
\textbf{FUNSD} \citep{funsd}, \textbf{CORD} \citep{cord},
\textbf{SROIE} \citep{sroie}, and \textbf{DocVQA} \citep{docvqa}.

\textbf{Granularity.} All datasets provide ground-truth annotations at the \textbf{word level}.
To simulate OCR post-processing under real-world conditions, we derive
\textbf{line-level} segments by merging vertically aligned word boxes.
This creates two granularity tiers-word and line-that reflect typical OCR system outputs. Table~\ref{tab:dataset_stats} shows the number of document images and
bounding boxes at both granularities.

\begin{table}[t]
  \centering\small
  \begin{tabular}{llccc}
    \hline
    \textbf{Dataset} & \textbf{Level} & \textbf{Train BBoxes} & \textbf{Test BBoxes} \\
    \hline
    FUNSD & Word & 21{,}888 & 8{,}707 \\
          & Line & 7{,}259  & 2{,}270 \\
    CORD  & Word & 19{,}370 & 2{,}356 \\
          & Line & 11{,}106 & 1{,}336 \\
    SROIE & Word & 73{,}747 & 40{,}411 \\
          & Line & 34{,}465 & 19{,}085 \\
    DocVQA & Word & 6,202,284 & 937,786 \\
           & Line & 1,806,853 & 259,582 \\
    \hline
  \end{tabular}
  \caption{Dataset statistics at word and derived line level}
  \label{tab:dataset_stats}
\end{table}

\noindent\textbf{Model and attack configuration.}
All VDU models are finetuned for 100 epochs with AdamW 
(learning rate $2{\times}10^{-5}$, batch 32, weight decay $1{\times}10^{-2}$) on a single NVIDIA L40S (48 GB).
Unless stated otherwise, we use text-edit budget $\rho=0.10$ and layout budgets $\tau\!\in\!\{0.9,0.75,0.6\}$.

%--------------------------------------------------

\subsection{Bounding Box Predictor Evaluation}
\label{ssec:bbx_pred_eval}

To enable PGD-based layout attacks, we train a lightweight 4-layer Transformer (\S\ref{ssec:bbx_pred}) that maps encoder embeddings to bounding box parameters \((c_x, c_y, \log w, \log h)\). Training targets are derived from clean ground-truth annotations at word or line level.

We train a bounding box predictor for each model separately, using the spatial embeddings extracted from that model when fine-tuned on each corresponding dataset.

\begin{table}[t]
  \centering\scriptsize
  \begin{tabular}{llcccc}
    \hline
    \textbf{Model} & \textbf{Gran.} & \textbf{FUNSD} & \textbf{CORD} & \textbf{SROIE} & \textbf{DocVQA} \\
    \hline
    LayoutLMv2     & Line & 71.78  & 73.29  & 70.23  & -- \\
    LayoutLMv2     & Word & 66.45  & 69.45  & 67.95     & -- \\
    LayoutLMv3     & Line & \textbf{89.34}  & \textbf{94.73}  & \textbf{94.05}     & -- \\
    LayoutLMv3     & Word & 84.55  & 93.17  & 91.41     & -- \\
    ERNIE          & Line & 70.84  & 86.09  & 88.71  & -- \\
    ERNIE          & Word & 72.84  & 80.74  & 88.95  & -- \\
    GeoLayoutLM    & Line & --  & --  & --     & -- \\
    GeoLayoutLM    & Word & --  & --  & --     & -- \\
    \hline
  \end{tabular}
  \caption{BBox prediction accuracy (mIoU, \%) across datasets and granularities. ``--'' indicates cases where no model was trained, or where PGD adversarial samples could not be generated due to predicted boxes falling below the budget threshold (IoU $< 0.6$).}
  \label{tab:bbox_miou}
\end{table}

Table~\ref{tab:bbox_miou} reports mIoU of the bounding box predictors used to generate PGD adversarial examples. Since LayoutLMv3 ~\citep{layoutlmv3} yields the most accurate bounding box predictions, we exclusively use its line-level bbox predictor to generate PGD adversarial samples across all evaluations. For other models, missing values (``--'') indicate either no training was performed or the predicted boxes do not meet the minimum IoU budget constraint for PGD attack (i.e., no adversarial box can be constructed with IoU $\ge$ 0.6).

%--------------------------------------------------

\subsection{Effectiveness of Budgeted Attacks}
\label{sec:effectiveness_attacks}

We comprehensively analyze how budget-controlled perturbations affect model robustness. We focus on six adversarial scenarios (S1--S6), and evaluate across attack methods (Random vs. PGD), granularities (word vs. line), and tasks (KIE vs. VQA). All attacks are conducted under a fixed budget (IoU $\geq$ 0.6), and results are averaged over 5 random seeds.

\subsubsection{Random Shift vs. PGD}

We evaluate Random Shift and PGD attacks across six scenarios at the \textbf{line level}, using LayoutLMv3 ~\citep{layoutlmv3} on four datasets.
\noindent According to Table~\ref{tab:line_vs_pgd_split}, PGD-based attacks consistently yield greater performance degradation than Random Shift in compound scenarios such as S5 and S6. This is attributable to the gradient-driven nature of PGD as well as the higher precision of the underlying bbox predictor. In our case, the BBox predictor from LayoutLMv3 embeddings achieves the highest mIoU, enabling more targeted and effective adversarial shifts. We observe similar trends on LayoutLMv2, ERNIE, and GeoLayoutLM, confirming the robustness and generality of these findings across architectures.

\begin{table*}[t]
  \centering
  \renewcommand{\arraystretch}{1.1}
  \setlength{\tabcolsep}{3.5pt}
  \scalebox{0.9}{\begin{tabular}{llcccccc|c}
    \hline
    & & \multicolumn{2}{c}{\begin{tabular}[c]{@{}c@{}}\textbf{FUNSD} \\ (F1 Drop \%)\end{tabular}} & 
        \multicolumn{2}{c}{\begin{tabular}[c]{@{}c@{}}\textbf{CORD} \\ (F1 Drop \%)\end{tabular}} & 
        \multicolumn{2}{c|}{\begin{tabular}[c]{@{}c@{}}\textbf{SROIE} \\ (F1 Drop \%)\end{tabular}} & 
        \begin{tabular}[c]{@{}c@{}}\textbf{DocVQA} \\ (ANLS Drop \%)\end{tabular} \\
    \cline{3-8}
    \textbf{Model} & \textbf{Scenario} & \textbf{Rand.} & \textbf{PGD} & \textbf{Rand.} & \textbf{PGD} & \textbf{Rand.} & \textbf{PGD} & \textbf{Rand. } \\
    \hline
    \multicolumn{9}{l}{\textit{\textbf{LayoutLMv3}}} \\
    & S1: BBox only               & 7.94  &  \textbf{13.32}    & 1.28  & \textbf{4.77}    & 0.46  & \textbf{5.59}    & 0.10  \\
    & S2: BBox + Pixel            & 16.24 &  \textbf{13.24}    & 3.14  & \textbf{4.99}    & 0.29  & \textbf{6.34}    & 0.01  \\
    & S3: S2 + Augment               & 17.80 &  \textbf{14.38}    & 3.14  & \textbf{5.24}    & 0.39  & \textbf{6.40}    & 0.08  \\
    & S4: Text only               & 7.31  & --        & 7.13  & --      & 20.78 & -- & 34.50 \\
    & S5: BBox + Text             & 16.55 & \textbf{22.78}     & 11.67 &  \textbf{11.80}       & 23.02 & \textbf{26.12} & \textbf{35.75} \\
    & S6: BBox + Pixel + Text     & 28.91 & \textbf{29.18} & 13.23 & \textbf{18.37} & \textbf{23.42} & \textbf{28.18} & 35.42 \\
  \end{tabular}}
  \caption{Drop in F$_1$ or ANLS (\%) under line-level attack for all six scenarios. For FUNSD, CORD, and SROIE, each pair of columns represents Random and PGD variants. Bold values indicate strongest degradation across settings. PGD values will be filled in separately if available.}
  \label{tab:line_vs_pgd_split}
\end{table*}

\subsubsection{Line vs. Word Granularity (LayoutLMv3)}

We compare line-level and word-level attacks on LayoutLMv3 ~\citep{layoutlmv3} across all scenarios using FUNSD ~\citep{funsd}. Table~\ref{tab:line_vs_word_gap} shows the gap between line and word F$_1$ drops by scenario and granularity:
\begin{table}[h]
  \centering
  \renewcommand{\arraystretch}{1.1}
  \setlength{\tabcolsep}{4pt}
  \scalebox{0.9}{
  \begin{tabular}{lcc}
    \hline
    \textbf{Scenario} & \textbf{Gap (Rand)} & \textbf{Gap (PGD)} \\
    \hline
    S1     & 7.60  & 11.98  \\
    S2     & 15.61 & 11.77  \\
    S3     & 15.93 & 13.29  \\
    S4     & 0.09  & --     \\
    S5     & 8.56  & 13.04  \\
    S6     & 21.37 & 13.44  \\
    \hline
  \end{tabular}}
  \caption{Gap in F$_1$ performance between line-level and word-level attacks on LayoutLMv3 (line drop minus word drop) under each scenario on FUNSD~\citep{funsd}. Positive values indicate stronger degradation from line-level attacks.}
  \label{tab:line_vs_word_gap}
\end{table}
\noindent Table~\ref{tab:line_vs_word_gap} reports the performance gap between line-level and word-level attacks on LayoutLMv3 across all scenarios. Positive values indicate that line-level attacks are more damaging. We observe that line-based perturbations consistently lead to larger F$_1$ drops than their word-level counterparts across all compound scenarios (S2–S6), for both Random and PGD. The largest gap appears in scenario S6 (layout + pixel + text), with a 21.4 pp difference under Random shift and 13.4 pp under PGD. This reflects how line boxes—being longer and semantically denser—induce broader misalignment, impacting multiple tokens and layout cues simultaneously.

\subsubsection{Cross-Task Robustness (KIE vs. VQA)}

Table~\ref{tab:line_vs_pgd_split} further compares the performance of LayoutLMv3 under line-level perturbation across KIE and VQA tasks. KIE datasets suffer more under layout-pixel shifts (S1--S3), while VQA exhibits greater degradation under text-related attacks (S4--S6). This divergence reflects task-specific dependencies: KIE relies on precise layout structure, while VQA depends more on accurate textual content.

%--------------------------------------------------
\subsection{Transferability of PGD Attacks}
\label{ssec:transferability}

% Giảm khoảng cách giữa các bảng
\setlength{\textfloatsep}{0pt}
\setlength{\floatsep}{0pt}
\setlength{\intextsep}{5pt}
\setlength{\dbltextfloatsep}{5pt}
\setlength{\dblfloatsep}{0pt}

We test the cross-model transferability of PGD attacks generated on LayoutLMv3 ~\citep{layoutlmv3}.

\subsubsection{Transfer to Text + Layout + Image models.}

These include LayoutLMv2 ~\citep{layoutlmv2}, ERNIE-Layout ~\citep{ernie}, and GeoLayoutLM ~\citep{geolayoutlm}. Despite incorporating visual encoders, these models still suffer significant F$_1$ drops under PGD transfer. FUNSD ~\cite{funsd} remains the most vulnerable dataset, particularly for LayoutLMv2 and GeoLayoutLM (55.5\% and 53.6\% drop respectively). ERNIE-Layout is the most robust overall, with all drops under 7\%, even under PGD.

\begin{table}[t]
  \centering
  \footnotesize
  \renewcommand{\arraystretch}{1.0}
  \setlength{\tabcolsep}{3pt} % Reset to default spacing
  \scalebox{0.9}{\begin{tabular}{p{2.5cm}ccc} % Only widen the 'Model' column
    \hline
    \textbf{Model} & \textbf{Sce.} &
    \begin{tabular}[c]{@{}c@{}}\textbf{Rand.} \\ \textbf{F1 Drop\%}\end{tabular} &
    \begin{tabular}[c]{@{}c@{}}\textbf{PGD Transfer} \\ \textbf{F1 Drop\%}\end{tabular} \\

    \hline
    \multicolumn{4}{l}{\textit{\textbf{LayoutLMv2}}} \\
    FUNSD & S1 & 18.72 & \textbf{20.21} \\
    FUNSD & S6 & 40.82 & \textbf{55.54} \\
    CORD  & S1 & 6.42 & \textbf{6.70} \\
    CORD  & S6 & 34.57 & \textbf{40.26} \\
    \multicolumn{4}{l}{\textit{\textbf{ERNIE-Layout}}} \\
    FUNSD & S1 & 5.03  & \textbf{7.50} \\
    FUNSD & S6 & 4.51  & \textbf{6.72} \\
    CORD  & S1 & 2.16  & \textbf{3.00} \\
    CORD  & S6 & 3.10  & \textbf{3.33} \\
    \multicolumn{4}{l}{\textit{\textbf{GeoLayoutLM *Word level}}} \\
    FUNSD & S1 & \textbf{1.14}  & 0.87 \\
    FUNSD & S6 & 49.08 & \textbf{53.56} \\
    CORD  & S1 & 0.04  & \textbf{0.58} \\
    CORD  & S6 & 10.44 & \textbf{12.75} \\
    \hline
  \end{tabular}}
  \caption{Transferability of LayoutLMv3-generated PGD examples to other \textit{text+layout+image} models under scenarios S1 and S6 (line-level). Each block corresponds to a model with dataset and scenario breakdown. Values are absolute F$_1$ drops (\%).}
  \label{tab:transfer_tli}
\end{table}

Table~\ref{tab:transfer_tli} shows PGD consistently causes higher performance degradation than Random Shift, particularly in scenario S6. This reinforces the effectiveness of gradient-based optimization when paired with accurate bbox predictors. In this setting, LayoutLMv3 provides the highest mIoU predictor, enabling highly effective transfer of perturbations. Meanwhile, ERNIE-Layout appears more robust, potentially due to architectural differences or reduced reliance on spatial features.

\subsubsection{Transfer to Text + Layout models}

\noindent Despite lacking visual encoders, models like LayTextLLM~\citep{laytextllm}, LLaMA3~\citep{llama3}, and ChatGPT 4.1 mini~\citep{openai_gpt41mini} still suffer non-trivial F$_1$ degradation under layout-based PGD attacks. Table~\ref{tab:transfer_tl} shows that PGD transfer leads to drops up to 3.4 pp on FUNSD and 7.9 pp on SROIE, while Random Shift causes negligible or even slightly negative effects. This confirms the strong cross-modality and cross-architecture transferability of PGD, even to models with no visual input or explicit layout supervision.

\begin{table}[t]
  \centering
  \renewcommand{\arraystretch}{1.0}
  \setlength{\tabcolsep}{3pt}
  \scalebox{0.8}{
  \begin{tabular}{ccc}
    \hline
    \textbf{Target Model} & 
    \begin{tabular}[c]{@{}c@{}}\textbf{Rand.} \\ \textbf{F1 Drop\%}\end{tabular} &
    \begin{tabular}[c]{@{}c@{}}\textbf{PGD Transfer} \\ \textbf{F1 Drop\%}\end{tabular} \\
    \hline
    LayTextLLM / FUNSD      & 0.73 & \textbf{2.87} \\
    LLaMA3-3B / FUNSD       & -1.12 & \textbf{2.50} \\
    ChatGPT 4.1 mini / FUNSD      & 1.92 & \textbf{3.37} \\
    LLaMA3-1B / FUNSD       & -0.17 & \textbf{0.77} \\
    \hline
    LayTextLLM / SROIE     & -0.05 & \textbf{0.93} \\
    LLaMA3-3B / SROIE      & 0.89 & \textbf{3.14} \\
    ChatGPT 4.1 mini / SROIE     & 4.34 & \textbf{6.83} \\
    LLaMA3-1B / SROIE      & 2.41 & \textbf{7.86} \\
    \hline
  \end{tabular}}
  \caption{Transferability of LayoutLMv3 ~\citep{layoutlmv3} PGD examples to \textit{text+layout} models under scenario S6 (line-level). PGD leads to greater degradation than Random shift, despite the lack of visual modality.}
  \label{tab:transfer_tl}
\end{table}

%--------------------------------------------------

\subsection{Ablation Study}

We conduct additional ablation studies to analyze the sensitivity of VDU models to three key factors: bounding box budget, adversarial transferability, and text modification strategies.

\noindent\textbf{Effect of Bounding Box Budget.} Table ~\ref{tab:bbox_budget} indicates that lowering the IoU threshold increases attack strength for both PGD and Random methods. Notably, PGD maintains higher effectiveness even under stricter constraints (e.g., 6.5\% drop at IoU 0.9 on FUNSD), while Random attacks quickly lose impact as the budget tightens (e.g., only 0.54\% drop). This highlights PGD's ability to find optimal perturbations within a tight search space.

\begin{table}[t]
  \centering
  \footnotesize
  \renewcommand{\arraystretch}{1.0}
  \setlength{\tabcolsep}{3pt}
  
  \begin{tabular}{lccc}
    \hline
    \textbf{Level} & \textbf{IoU Budget} & \textbf{Attack} & 
    \begin{tabular}[c]{@{}c@{}}\textbf{F1 Drop\%} \\ (FUNSD / CORD)\end{tabular} \\
    \hline
    Line & 0.6 & Random & 7.94 / 1.28 \\
    Line & 0.75 & Random & 2.94 / 0.22 \\
    Line & 0.9 & Random & 0.54 / 0.15 \\
    Line & 0.6 & PGD & 13.32 / 4.77 \\
    Line & 0.75 & PGD & 6.60 / 0.92 \\
    Line & 0.9 & PGD & 6.50 / 0.51 \\
    \hline
  \end{tabular}
  \caption{F$_1$ drops under BBox attacks (scenario S1), across different IoU budgets and attack methods. PGD remains more effective under tight constraints.}
  \label{tab:bbox_budget}
\end{table}

\noindent\textbf{Text Modification Strategies.} We compare two text-only attack methods: (1) random character replacement with 10\% edit rate, and (2) a Unicode diacritic attack following \citep{whenvisionfails}, which uses visually confusable glyphs via combining diacritical marks. Results in Table~\ref{tab:textmod} show that Unicode attacks consistently cause greater performance degradation-up to 22.4\% on CORD-highlighting their stronger disruption of semantic and visual consistency.
\begin{table}[t]
  \centering
  % \scriptsize
  \scalebox{0.7}{\begin{tabular}{llcc}
    \hline
    \textbf{Model} & \textbf{Attack} & \textbf{FUNSD Drop\%} & \textbf{CORD Drop\%} \\
    \hline
    LayoutLMv3 & S3 text only 0.1  & 7.31 & 7.13 \\
    LayoutLMv3 & Unicode diacritic & 16.75 & 22.35 \\
    \hline
  \end{tabular}}
  \caption{Text-only attacks: Unicode-based modifications cause larger degradation compared to random character replacement.}
  \label{tab:textmod}
\end{table}

%=====================================================================

\section{Conclusion}
We present a unified adversarial framework for evaluating the robustness of OCR-based Visual Document Understanding (VDU) systems across layout, pixel, and text modalities. Our method incorporates attack granularity (word vs. line) and budget constraints to simulate realistic perturbations.

Our contributions include: (1) a budgeted attack strategy combining layout, pixel, and text; (2) a differentiable BBox predictor enabling gradient-based layout attacks despite discrete spatial encodings; (3) a six-scenario benchmark over four datasets; and (4) in-depth analysis of modality combinations, granularity effects, and transferability.

Experiments reveal major vulnerabilities in state-of-the-art VDU models. Layout perturbations (S1) degrade performance significantly, with compound attacks (S6) amplifying this effect. PGD-based layout attacks consistently outperform random shift, particularly at tighter IoU budgets. Line-level attacks are more destructive than word-level due to broader context impact, and Unicode diacritic text attacks show stronger disruption than random edits. PGD samples also transfer well across models, including to those without image inputs.

\section*{Limitations.} Our work focuses on OCR-based systems; models like LayTextLLM ~\citep{laytextllm} exhibit resilience due to shuffled box encoding. We also do not evaluate OCR-free models (e.g., Qwen-VL ~\citep{qwen25vltechnicalreport}, DeepSeek-VL ~\citep{deepseekvl}), which rely on pure visual grounding and may require different attack strategies. Furthermore, our analysis is limited to white-box access; extending to black-box settings where only model outputs are observable remains future work.

\section*{Ethical Considerations}

Our study aims to evaluate the robustness of Visual Document Understanding systems by exposing vulnerabilities through controlled adversarial perturbations. All attacks are simulated under strict budget constraints to reflect plausible real-world issues. The adversarial examples are generated for research purposes only and are not intended for malicious use. No sensitive personal data are used or exposed in our experiments.

% Bibliography entries for the entire Anthology, followed by custom entries
%\bibliography{anthology,custom}
% Custom bibliography entries only
\bibliographystyle{acl_natbib}
\bibliography{custom}

% \appendix

% \section{Example Appendix}
% \label{sec:appendix}

% This is an appendix.

\end{document}